\documentclass[runningheads]{llncs}
\usepackage{graphicx}
\usepackage{hyperref}
\usepackage[misc]{ifsym}
\begin{document}
\title{Speaker Recognition using SincNet and X-Vector Fusion}
\author{Mayank Tripathi \and
Divyanshu Singh \and
Seba Susan \textsuperscript{(\Letter)} \orcidID{0000-0002-6709-6591} }
\authorrunning{Tripathi, Singh and Susan}

\institute{Department of Information Technology \\ Delhi Technological University, Delhi 110042, India \\
\email{\{mayank\_bt2k16,divyanshu\_bt2k16\}@dtu.ac.in,  seba\_406@yahoo.in}}
\maketitle              % typeset the header of the contribution
\begin{abstract}
 In this paper we propose an innovative approach to perform speaker recognition by fusing two recently introduced deep neural networks (DNNs) namely - SincNet and X-Vector. The idea behind using SincNet filters on the raw speech waveform is to extract more distinguishing frequency-related features in the initial convolution layers of the CNN architecture. X-Vectors are used to take advantage of the fact that this embedding is an efficient method to churn out fixed dimension features from variable length speech utterances, something which is challenging in plain CNN techniques, making it efficient both in terms of speed and accuracy. Our approach uses the best of both worlds by combining X- vector in the later layers while using SincNet filters in the initial layers of our deep model. This approach allows the network to learn better embedding and converge quicker. Previous works use either X-Vector or SincNet Filters or some modifications, however we introduce a novel  fusion architecture wherein we have combined both the techniques to gather more information about the speech signal hence, giving us better results. Our method focuses on the VoxCeleb1 dataset for speaker recognition, and we have used it for both training and testing purposes. 
\end{abstract}
\section{Introduction}
Speaker Recognition and Automatic Speech Recognition(ASR) are two of the actively researched and interesting fields in the computer science domain. Speaker recognition has applications in various fields such as biometric authentication, forensics, security and speech recognition, which has contributed to steady interest in this discipline \cite{beigi2011speaker}. The conventional method of speaker identification involves classification of features extracted from speech such as the Mel Frequency Cepstral Coefficients (MFCC) \cite{susan2012fuzzy}. 

With the advent of i-vectors \cite{dehak2010front}, speaker verification has become faster and more efficient as compared to the preceding model based on the higher dimensional Gaussian Mixture Model (GMM) supervectors. The i-vectors are low dimensional vectors that are rich in information and represent distinguishing features of the speaker. i-vectors are used to generate a verification score between the embedding of two speakers. This score gives us information about whether both are the same speaker or different speakers. Previous experiments have shown that i-vectors perform better with Probabilistic Linear Discriminant Analysis (PLDA) \cite{kenny2010bayesian}. Work has also been carried  out  in  the  field  of  training the i-vectors using different techniques in order to get better embedding, therefore, better results \cite{ghalehjegh2015deep,novotny2019discriminatively}.

Currently, researchers are moving towards Deep Neural Network (DNN) to obtain speaker embedding. DNN can be directly optimized to distinguish between speakers. DNN showed promising results in comparison to statistical measures such as i-vectors \cite{huang2015investigation,snyder2017deep}. X-vectors are seen as an improvement over the i-vector system (which is also a fixed-dimension vector system) because they are more robust and have yielded better results \cite{snyder2018x}. X-vector extraction methodology employs a Time-Delayed Neural Network (TDNN) to learn features from the variable-length audio samples and converts them to fixed dimension vectors. This architecture can broadly be broken down in their order of occurrence into three units, namely, frame-level layers, statistics pooling layer and the segment level layers. X- vectors can then be used with classifiers of any kind to carry out recognition tasks.

SincNet \cite{ravanelli2018speaker} is a deep neural network that has embedded band pass filters for extracting features from the audio sample. The features are then fed into DNN based classifiers. We have used SincNet filters and not fed the audio waveform directly into the DNN based classifiers as the latter technique poses problems like high convergence time and less appealing results. The SincNet filters are actually band pass filters which are derived from parameterized sinc functions. This gives us a compact and an efficient way to get a filter bank that can be customized and specifically tuned for a given application.

Our novel architecture combines the goodness of both the methodologies - SincNet and X-Vector, by extracting features using both the techniques and feeding them to fully connected dense layers which acts as a classifier. The organization of this paper is as follows: the related works are described in section 2, the proposed fusion architecture is presented in section 3, the experimental setup and the results are discussed in sections 4 and 5 respectively, and the final conclusions are drawn in section 5. 

\section{Related Works}
The i-vectors \cite{dehak2010front} feature extraction method has proved to be state-of-the-art for quite some time now for the speaker recognition tasks. Techniques like the Probabilistic Linear Discriminant Analysis (PLDA) and Gauss-PLDA \cite{cumani2013probabilistic,kenny2010bayesian} are fed with the extracted features and based upon which they carry out classification. Despite being a state-of-the-art technique, we can still improve the performance in terms of accuracy.

With the advent of various deep learning techniques in multiple domains for feature extraction, work has also been carried out to extract features from audio signals using deep learning architectures \cite{li2017deep}. These architectures can range from using deep learning just for feature extraction to using neural networks as classifiers as well. The deep learning methods give better results when compared to older techniques based on feature engineering \cite{variani2014deep}. 

The most commonly used deep learning method for feature extraction is the one based on the Convolution Neural Network (CNN) architecture \cite{snyder2017deep}. The CNN has been a preferred choice for researchers as it has given good quality results in tasks such as image recognition. Initially the CNN was fed with spectrogram in order to extract features from the audio signals \cite{chung2018voxceleb2,palaz2015analysis,sainath2015learning,snyder2017deep,zhang2018text}. Despite the fact that, spectrogram-based CNN methods were giving good results there were again many drawbacks of using this method \cite{wyse2017audio}. Firstly, the spectrogram is the temporal representation of data unlike images which are spatio-temporal representations. Secondly, a frame in spectrogram can be a result of superposition of multiple waves and not just a single wave which means that a single result can be obtained using different waves super-positioned in different manner. Also, even though spectrograms retain more information than standard hand-crafted features, their design still requires careful tuning of some hyper-parameters, such as the duration, overlap, and typology of the frame window.

The above drawbacks also inspired researchers to directly input raw waveforms into CNN \cite{lee2017raw} so that no information is lost. This was a good methodology to follow but it results in slower convergence since it processes the complete frequency band of the audio signal.

SincNet \cite{ravanelli2018speaker} and X-vector \cite{snyder2018x} are amongst the most recent deep learning methods for speech signal classification. Both of them have proved to be more robust than methods which have preceded them. 

\section{Proposed Fusion Model}
We propose a novel fusion of SincNet \cite{ravanelli2018speaker} and X-Vector \cite{snyder2018x}  embedding which will enable us to take the temporal features into account which is quite important for any audio based recognition task since the signal at any point at time \textit{t} in an audio signal is affected by points preceding and succeeding it. 
\subsection{X-Vector Embedding}
We have used a pre-trained X-vector system which was trained on the VoxCeleb1 \cite{nagrani2017voxceleb}  dataset which we are using. The pre-trained x-vector system is available in the kaldi toolkit which is available for public use \cite{kaldi}. Table.~\ref{tab1} shows the architecture of the x-vector feature extractor system which has been trained on the VoxCeleb1 dataset. X-vector extraction methodology employs a Time-Delayed Neural Network (TDNN) to learn features from the variable-length audio samples and converts them to fixed dimension vectors. This architecture can broadly be broken down in their order of occurrence into three units, namely, frame-level layers, statistics pooling layer and the segment level layers. X- vectors can then be used with classifiers of any kind to carry out recognition tasks.

\begin{table}
\centering
\caption{X-vector DNN architecture \cite{snyder2018x}}\label{tab1}
\begin{tabular}{l|l|l|l}
\hline
Layer &  Layer Context & Table Context & Input x Output\\
\hline
frame 1 &  [t-2, t+2] & 5 & 120x512\\
frame 2 &  \{t-2, t, t+2\} & 9 & 1536x512\\
frame 3 &  \{t-3, t, t+3\} & 15 & 1536x512\\
frame 4 &  \{t\} & 15 & 512x512\\
frame 5 &  \{t\} & 15 & 512x1500\\
stats pooling &  [0, T) & T & 1500Tx3000\\
segment 6 &  \{0\} & T & 3000x512\\
segment 7 &  \{0\} & T & 512x512\\
softmax &  \{0\} & T & 512xN\\
\hline
\end{tabular}
\end{table}

\subsection{SincNet Architecture}
The SincNet architecture implements various band pass filters in its initial layers, that learns from raw audio signals. The SincNet filters are implemented using a set of mathematical operations which are stated below \cite{ravanelli2018speaker}.

\noindent The function $h[.]$ is the Finite Impulse Response (FIR) filter used as a convolution filter.
\begin{equation}
y[ n] \ =\ x[ n] \ *\ h[ n] \ =\ \sum ^{L-1}_{l=0} \ \ x[ l] *\ h[ n\ -\ l]
\end{equation}

\noindent $g[.]$ is a predefined function that depends on a few learnable parameters $\theta$.
\begin{equation}
y[ n] \ =\ x[ n] \ *\ g[ n,\theta ]
\end{equation}

\noindent$rect(.)$  is the rectangular function in the magnitude frequency domain and $f_{1}$ and  $f_{2}$ are cut-off frequencies.
\begin{equation}
G[ f,f_{1} ,f_{2}] \ =\ rect\left(\frac{f}{2f_{2}}\right) \ \ -\ rect\left(\frac{f}{2f_{1}}\right)
\end{equation}
\noindent Here, the $sinc$  function is defined as $sinc( x) \ =\ \sin( x) \ /\ x$   
\begin{equation}
g[ n,f_{1} ,f_{2}] \ =\ 2f_{2} sinc( 2\pi f_{2} n) \ -\ 2f_{1} sinc( 2\pi f_{1} n)
\end{equation}
\noindent We have to ensure that $f_{1} \geq 0$  and $f_{2} \geq f_{1}$  , therefore, the previous equation is actually fed with the following parameters: 
\begin{equation}
f^{abs}_{1} \ \ =\ |f1|\\
\end{equation}
\begin{equation}
f^{abs}_{2} \ \ =\ f1\ +\ |f2\ -\ f1|
\end{equation}
\noindent Windowing is performed by multiplying the function  with a window function $w$.
\begin{equation}
g_{w}[ n,f_{1} ,f_{2}] =\ g[ n,f_{1} ,f_{2}] *\ w[ n]
\end{equation}
\noindent The windowing function $w$ is a hamming window which is given by $eqn(8)$.
 \begin{equation}
w[ n] =\ 0.54\ -\ 0.46*\ cos\left(\frac{2\pi n}{L}\right)
\end{equation}

\begin{figure}
\centering
\includegraphics[width=50mm]{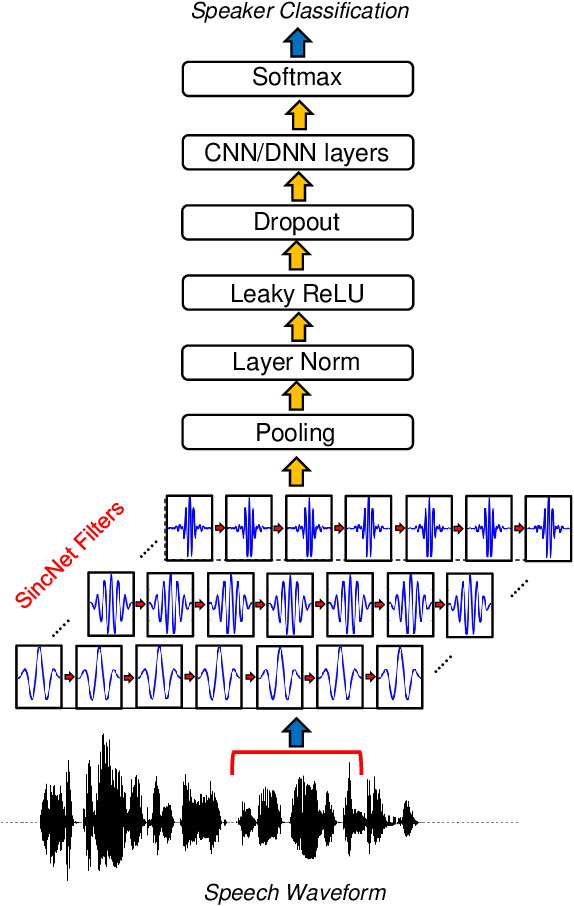}
\caption{This is the SincNet architecture. The image has been taken from the original paper \cite{ravanelli2018speaker}.} \label{fig1}
\end{figure}

Once the filters are applied on the raw audio, we get features. These features can now be fed into any classifier. The vanilla SincNet architecture can be seen in Fig.~\ref{fig1}. It takes raw audio signals as input, applies SincNet filters and then feeds it into a CNN model which is used as a classifier.

\subsection{Fusion Model}
Our proposed fusion model fuses the pre-trained X-Vector with features extracted from the trained SincNet model. The concatenated features are fed to a fully connected dense layer. This is followed by two more fully connected dense layers as shown in Fig.~\ref{fig2}. The idea behind using SincNet filters on the raw waveform is to extract more distinguishing features in the initial convolution layers of the CNN architecture. X-Vectors are used to take advantage of the fact that this embedding is an efficient method to churn out fixed dimension features variable length speech utterances, something which is challenging in plain CNN techniques, making it efficient both in terms of speed and accuracy.

\section{Experimental Setup}
\subsection{Dataset}
We have carried out our experiments on the publicly available dataset VoxCeleb1 \cite{nagrani2017voxceleb}. VoxCeleb is an audio-visual dataset consisting of short clips of human speech, extracted from interview videos uploaded to YouTube. We have used the raw audio files for our experiments.
 The VoxCeleb1 dataset consists of videos from 1,251 celebrity speakers. This means a size of 1,251 speakers and about 21k recordings.

\begin{table}
\centering
\caption{VoxCeleb1 dataset distribution \cite{nagrani2017voxceleb}.}\label{tab2}
\begin{tabular}{l|l|l}
\hline
 &  Dev & Test\\
\hline
Number of speakers  &  1,251 & 1,251\\
Number of videos &  21,245 & 1,251\\
Number of utterances  &  145,265 & 8,251\\
\hline
\end{tabular}
\end{table}
 
\subsection{Model Architecture}
\begin{figure}
\centering
\includegraphics[width=94mm]{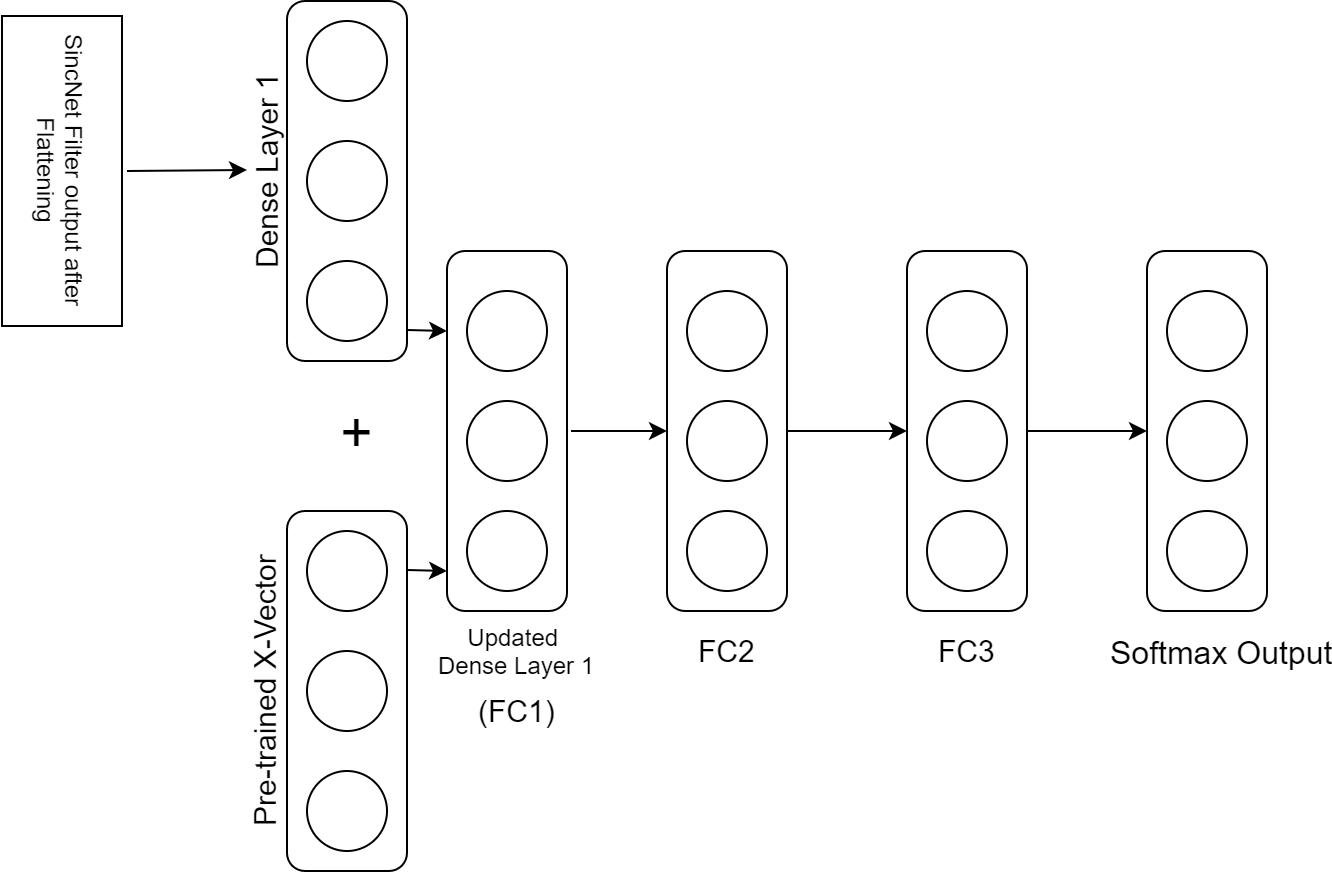}
\caption{The proposed fusion model.} \label{fig2}
\end{figure}

In order to carry out comparative results we have experimented with the original SincNet architecture and our proposed architecture. The original SincNet architecture uses SincNet filters for feature extraction whereas our architecture makes use of both SincNet and X-Vector. The classifier used in all the cases consists of several fully connected layers (DNN classifier) with softmax layer as the output layer. The models can be categorized as:
\begin{enumerate}
    \item SincNet based feature extractor and DNN classifier.
    \item X-Vector embedding and DNN classifier.
    \item X-Vector and SincNet based feature extractor and DNN classifier \textit{(Proposed)}.
\end{enumerate}
Our proposed fusion model fuses the pre-trained X-Vector with features extracted from the trained SincNet model. The concatenated features are fed to a fully connected dense layers as shown in Fig.~\ref{fig2}.The output obtained after convolution step is flattened and fed into the Dense Layer which is further concatenated with X-vector Embedding. The X-vector combined with the Dense layer constitutes fully connected layer 1 (FC1) which is further connected to FC2 and FC3. All the dense layers use Leaky ReLU as activation. The softmax layer is used as the output layer.

%
% the environments 'definition', 'lemma', 'proposition', 'corollary',
% 'remark', and 'example' are defined in the LLNCS documentclass as well.
%
\section{Experimental Results}
The experiments were carried out in Python 3 on a Nvidia \textit{GeForce GTX 1080 Ti
} GPU equipped with 16GB of memory. The software implementation of our code is made available online on \href{https://github.com/mayank408/ICAISC-Speaker-Identification-System}{GitHub} \cite{code_implementation}.  
The tests were carried out on the VoxCeleb1 dataset and the results are summarised in Table.~\ref{tab3}.  We calculated the Equal Error Rate (EER) for all the experiments and lower the value the of EER, the better is our model. The results were in alignment with our expectations which means that our system performed the best out of all the architectures tested.

\begin{figure}
\centering
\includegraphics[width=98mm]{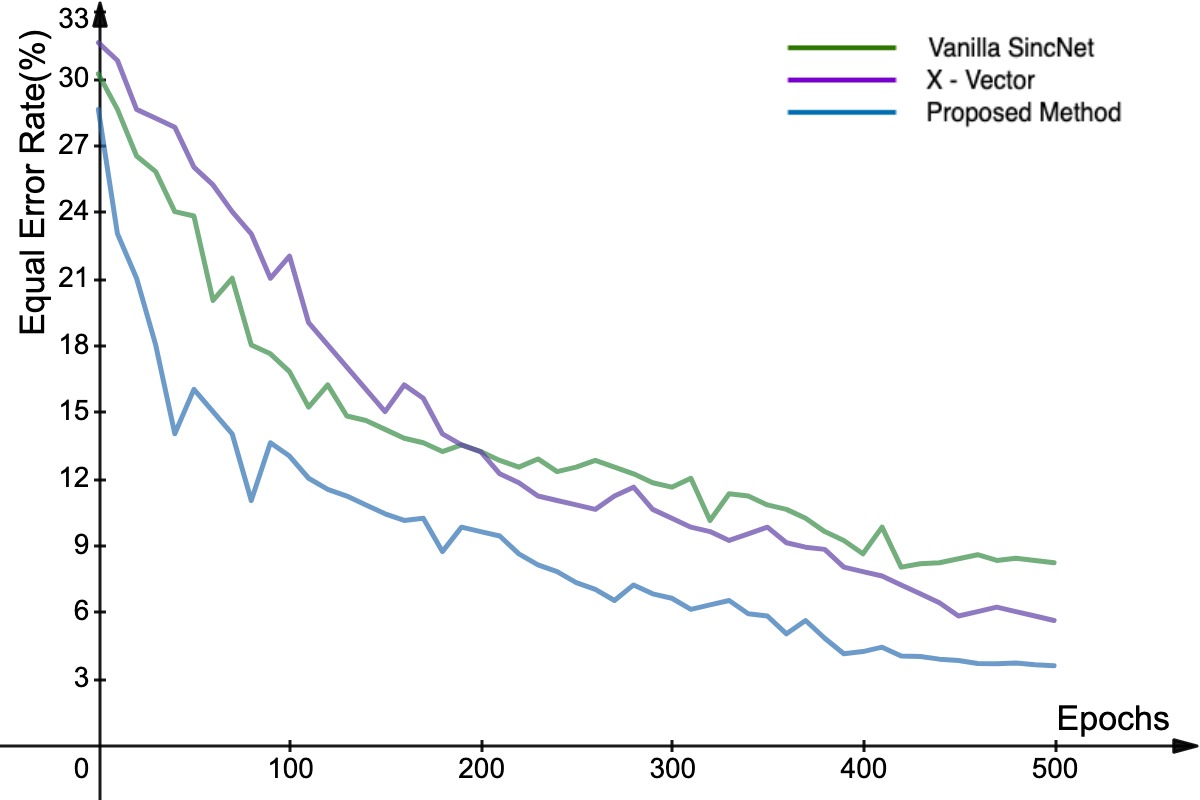}
\caption{Comparison of EER score of various architectures over epochs.} \label{fig3}
\end{figure}

We have an EER score 8.2 for pure SincNet based architecture, an EER score of 5.8 for the X-vector based architecture, and the best EER score of 3.56 using our SincNet and X-vector embedding based architecture. Our architecture proposed a framework which resulted in the EER score of 3.56 which was an improvement over the previous best EER score of 4.16 using X-vector over SITW core \cite{snyder2018x}. Fig.~\ref{fig3} shows a comparison of EER score of various architectures as the number of epochs increases. The proposed fusion model exhibits a consistently low EER over all epochs.

\begin{table}
\centering
\caption{Experimental Results}\label{tab3}
\begin{tabular}{l|l|l}
\hline
Architecture Used &  Training Dataset & EER (On Test Data)\\
\hline
SincNet  &  VoxCeleb1 & 8.2\\
X-Vector &  VoxCeleb1 & 5.8\\
\bfseries Proposed &  \bfseries VoxCeleb1 & \bfseries 3.56\\
\hline
\end{tabular}
\end{table}

\section{Conclusions}
In this paper we propose a novel fusion model involving two successful deep architectures for speaker recognition:- SincNet and X-Vector.The features extracted from the two sources are fused by concatenation and learnt using fully connected dense layers. We achieved an increase in the EER score from current state-of-the-art by 14.5\%\ on the VoxCeleb1 Dataset.
  It also showed quite a significant improvement over using vanilla SincNet which resulted in an EER score of 8.2.
Further improvement over this architecture can be carried out by combining the VoxCeleb2 dataset along with VoxCeleb1 dataset and using noise removal techniques prior to feeding into the network.

\bibliographystyle{splncs04}
\bibliography{bibliography}
\end{document}